\documentclass[conference]{IEEEtran}
\usepackage{times}
\usepackage{xcolor}
\usepackage{graphicx}
\usepackage{cite}
\usepackage{multicol}
\usepackage[bookmarks=true]{hyperref}
\usepackage{booktabs}  
\usepackage{multirow}
\usepackage{colortbl}  
\usepackage{caption}  
\usepackage{pifont}  
\usepackage{amsmath}
\usepackage{amssymb}
\usepackage{bm}
\usepackage{makecell}

\begin{document}

\newcommand{\name}{ViPSim}

\title{ViPSim: Collaborating Visual and Parameter Spaces for Consistent Long-Horizon Embodied World Models}

\author{\authorblockN{
Longyu Chen$^{1, \dagger}$,
Heng Li$^{1,2, \dagger}$,
Wei Yang$^{2}$,
Manqi Zhao$^{1}$\textsuperscript{*}
and
Dongsheng Jiang$^{1}$\textsuperscript{*}
}
\smallskip
\authorblockA{
$^{1}$Huawei Technologies Co., Ltd. $^{2}$Huazhong University of Science and Technology \\
$^{\dagger}$Equal Contribution, \textsuperscript{*}Corresponding author
}
}

\twocolumn[{%
\renewcommand\twocolumn[1][]{#1}%
\maketitle
}]

\begin{abstract}

Embodied World Models (EWMs) have emerged as a scalable and risk-free paradigm for advancing embodied intelligence, enabling the safety-critical evaluation of Vision-Language-Action systems. However, their reliability as evaluation benchmarks and foundational simulators is often hindered by the representation gap between low-dimensional actions and high-dimensional video synthesis. This gap results in a lack of geometric correspondence, manifesting as accumulated trajectory drift and inconsistent robot-object interactions during long-horizon rollouts. To bridge this gap, we propose \name, a framework that achieves consistent long-horizon generation through the synergistic collaboration of Visual and Parameter Spaces. We define the Visual Space as a domain of explicit spatial priors, integrating pixel-aligned projections of end-effector pose, camera perspectives, depth-informed scene geometry, and robotic morphological masks to provide dense structural grounding. Concurrently, the Parameter Space serves as a domain of numerical drivers, injecting raw action sequences and camera matrices to provide precise motion guidance. By unifying these two spaces, \name\ ensures that the generated states are simultaneously anchored by geometric boundaries and steered by numerical commands. Extensive experiments demonstrate that \name\ is backbone-agnostic and significantly enhances trajectory consistency. Notably, our approach exhibits emergent capabilities in generating complex interactions with deformable objects (e.g., cloth folding) and maintains robust performance in out-of-distribution and cross-embodiment scenarios, providing a high-fidelity foundation for the automated evaluation and predictive control of embodied agents.
\end{abstract}

\IEEEpeerreviewmaketitle

\section{Introduction}

The rapid evolution of Embodied AI has catalyzed the development of general-purpose robots, with Vision-Language-Action (VLA) models~\cite{pi0.5, DepthVLA, Evo, GeminiRobotics, UniVLA, GigaBrain} emerging as the cognitive core bridging perception and execution. However, the scarcity of efficient and reliable evaluation frameworks remains a critical bottleneck for the scaling of VLA systems. Traditional real-world evaluation is prohibitively expensive and carries inherent safety risks, while procedural rendering engines often struggle to accurately model the visual complexity of the real world, particularly those involving non-rigid interactions and deformable objects, leading to a significant domain gap.

Recently, generative Embodied World Models (EWMs)~\cite{Vid2World, WorldGym, WorldEval, MTVworld, ENACT, Gen2Act} have gained prominence as high-fidelity alternatives for evaluating robot policies. By predicting future visual states conditioned on current observations and actions, EWMs serve as data-driven benchmarks that offer high-throughput, risk-free environments—enabling the exploration of edge-case scenarios without the threat of hardware damage or hazardous physical collisions. While particularly salient for VLA evaluation, the impact of high-fidelity EWMs extends beyond benchmarking; they hold the potential to function as foundational simulators for model-based reinforcement learning, active inference, and predictive control pipelines, significantly accelerating the development cycle of embodied agents.

\begin{figure}[t]
    \centering
    \includegraphics[width=\linewidth]{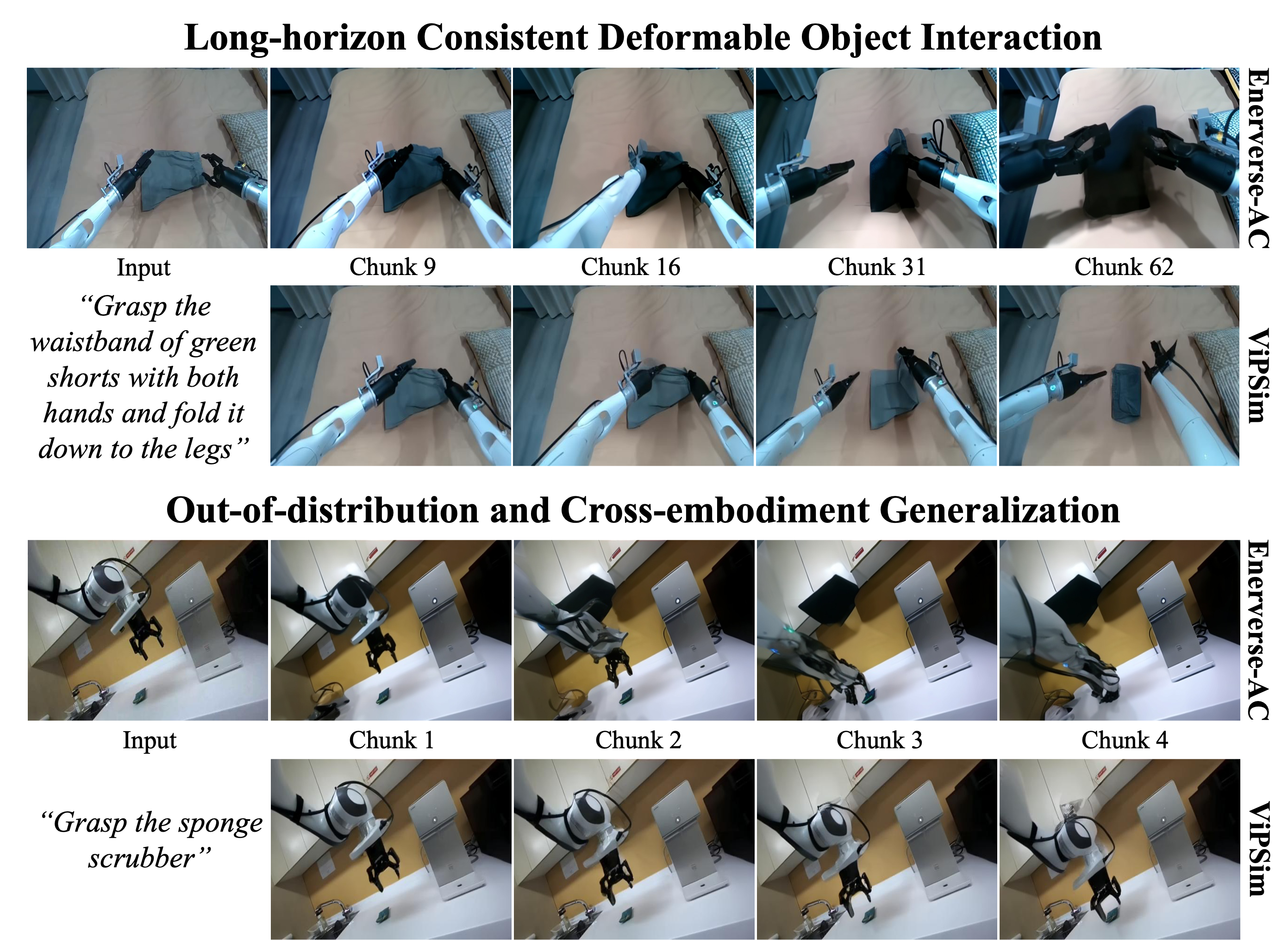}
    \vspace*{-5mm}
    \caption{Qualitative results on diverse tasks. The first sequence demonstrates long-horizon consistency in non-rigid object manipulation, where \name\ suppresses cumulative errors and maintains structural integrity. The second sequence highlights emergent cross-embodiment generalization: trained solely on Agibot, the model accurately emulates the Droid actions in OOD environments, establishing a reliable foundation for embodied evaluation. Extended zero-shot cross-domain results are provided in Appendix C.}
    \vspace{-6mm}
    \label{fig:main_exp}
\end{figure}

Despite their potential, existing action-conditioned EWMs\cite{EnerVerse, Genie-Envisioner, Cosmos, Ctrl-World, wang2025learningrealworldactionvideodynamics, UWM, unifolm-wma-0, WristWorld, MagicWorld} face significant challenges in maintaining consistency over long-horizon rollouts. We identify the root cause as a representation gap between the high-dimensional visual synthesis and the low-dimensional parameter inputs. Current architectures often fail to establish a rigorous geometric correspondence between numerical actions and the pixel-level scene dynamics. This lack of structural collaboration leads to severe cumulative trajectory drift. While recent efforts~\cite{RoboScape, PEWM, TesserAct, ORV, TraceGen} have introduced auxiliary visual cues, they often focus on prioritizing global scene coherence without collaborating explicit spatial specifications with numerical guidance, resulting in a fidelity mismatch in high-contact regions where robot morphology and scene geometry are most tightly coupled.

To bridge this gap, we present \name, a framework designed for consistent long-horizon EWM through the synergistic collaboration of Visual and Parameter Spaces. Specifically, the Visual Space establishes explicit spatial and morphological grounding. By projecting actions and camera perspectives into the image domain—while incorporating depth-informed scene geometry and robot-specific morphological masks—it creates a unified, dense representation. This provides the generative backbone with dense geometric priors, explicitly defining the spatial boundaries of interactions and the structural appearance of the embodiment. Complementary to this, the Parameter Space facilitates numerical command injection. It integrates action sequences and camera matrices into the generative backbone, providing the precise numerical guidance that dictates the quantitative unfolding of motion. This dual-space collaboration ensures that the generated visual flow is simultaneously anchored by spatial boundaries and steered by precise numerical commands. By establishing this rigorous correspondence, \name\ effectively suppresses cumulative trajectory drift and maintains high-fidelity consistency over extended rollouts, as visualized in \autoref{fig:main_exp}. Our core contributions are summarized as follows:

\begin{itemize}
     \item We present \name, an embodied world model that bridges the representation gap between low-dimensional actions and high-dimensional visual synthesis. \name\ achieves drift-free, long-horizon generation and maintains high consistency even in complex interactions with deformable objects.
     \item We introduce a dual-space collaboration mechanism that synchronizes geometric, morphological, and numerical priors by anchoring them across complementary domains. This mechanism ensures generated sequences adhere to both explicit spatial grounding (Visual Space) and precise numerical guidance (Parameter Space). 
     \item We provide comprehensive evidence that \name\ consistently improves trajectory consistency across diverse generative backbones. Furthermore, we demonstrate its robust generalization in out-of-distribution scenarios and cross-embodiment tasks, providing a high-fidelity and scalable foundation for the automated evaluation of embodied agents.
\end{itemize}

\section{Related work}
\subsection{Generative World Models}

Driven by the scaling of UNet-based~\cite{dynamicrafter} and DiT-based~\cite{WAN} architectures, generative models have achieved high-fidelity synthesis with significant temporal consistency. Several recent works extend generative models to interactive world models ~\cite{genie}, which aim to predict future states conditioned on current observations and actions. Transitioning these capabilities to Embodied AI, iVideoGPT~\cite{ivideogpt} integrates proprioception into sequence modeling. While these models excel in visual quality, they often lack the rigorous structural grounding required for precise robotic manipulation.

\subsection{Action-Conditioned Embodied World Models}

To ensure alignment between control and synthesis, prior works have explored various action-conditioned embodied world models. IRASim~\cite{IRASim} and Ctrl-World~\cite{Ctrl-World} utilize frame-level conditioning and memory retrieval to improve interactive consistency. However, directly injecting raw actions often fails to bridge the gap between low-dimensional action and high-dimensional pixels. To address this, EnerVerse-AC~\cite{EnerVerse-AC} introduced projected action maps to enhance generalization. In this work, we carefully design the action-injection pathway through a dedicated Parameter Space, ensuring long-term stability and cross-embodiment robustness.

Distinct from recent unified world models (UWM)~\cite{UWM} that integrate policy and world modeling into a single Transformer for multitask scaling, \name\ proposes a dual-space collaboration paradigm. Specifically, our model utilizes a pixel-aligned Visual Space to anchor object locations and a high-precision Parameter Space to steer motion. This design ensures long-term stability and cross-embodiment robustness, while we position \name\ as a precision-centric complement to scaling-centric frameworks like UWM: while the latter focuses on scaling policy learning, \name\ delivers high-fidelity, long-horizon visual rollouts essential for safe model-based evaluation and trajectory verification.

\subsection{Visual-Enhanced Embodied World Model}

Another research thread focuses on enhancing structural fidelity of EWMs via geometric priors. TesserAct~\cite{TesserAct} introduces a text-driven 4D EWM using RGB-DN data to bridge textual semantics and embodied execution. RoboScape~\cite{RoboScape} and ORV~\cite{ORV} utilize adaptive keypoint tracking and 4D semantic occupancy to reinforce 3D consistency, while PEWM~\cite{PEWM} adopts a hierarchical structure centered on action primitives to enhance skill reusability. Although these methods improve scene-level coherence, they often struggle in local high-contact regions where robot-object interactions are most intense. In contrast, our work collaborates visual enhancements with precise parametric guidance to ensure consistency in complex, long-horizon tasks.

\section{Method}

\subsection{Problem Formulation and Overview}

\begin{figure}[t] 
    \centering
    \includegraphics[width=1.0\linewidth]{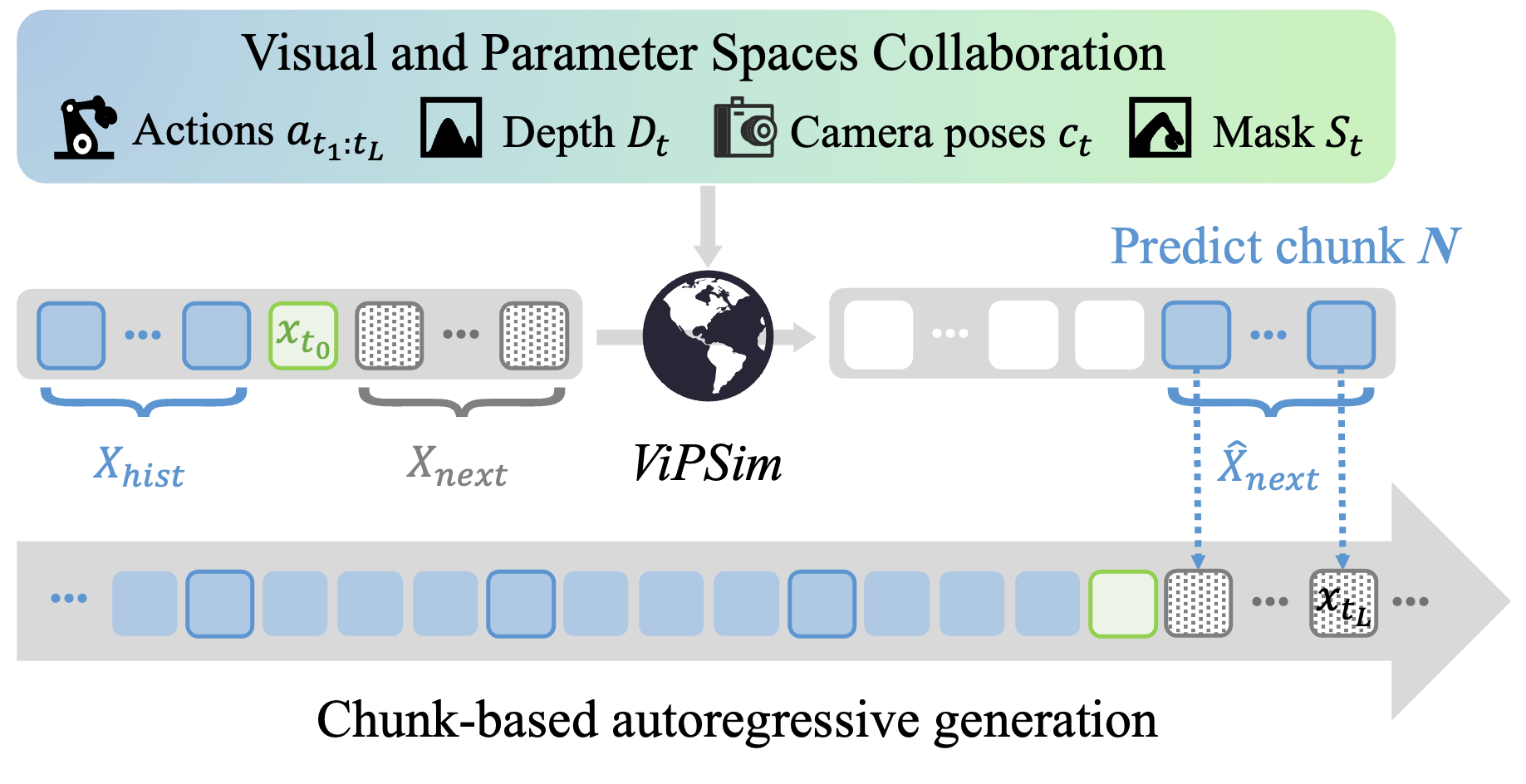} 
    \vspace{-5mm}
    \caption{Chunk-based Autoregressive Generation Pipeline.}
    \vspace{-6mm}
    \label{fig:ARgeneration}
\end{figure}

\noindent \textbf{Problem Formulation.} We build our framework upon video diffusion models \cite{dynamicrafter, WAN}, formulating the embodied generation task as a chunk-based autoregressive process. Given the high-dimensional nature of long-horizon robot videos, we decompose the generation into discrete chunks of length $L$. At each step, the model predicts the subsequent $L$ future frames $\hat{\mathbf{X}}_{next} = \{x_{t_1}, \dots, x_{t_L}\}$ conditioned on a multi-scale context. Specifically, the historical context consists of a current state frame $x_{t_0}$ and a set of $N_{hist}$ sparse historical frames $\mathbf{X}_{hist} = \{x_{\tau_1}, \dots, x_{\tau_{N_{hist}}}\}$ sampled from the preceding sequence ($\tau_1 < \dots < \tau_{N_{hist}} < t_0$), providing both instantaneous appearance anchors and long-range temporal consistency. The future control sequences include camera parameters $\mathbf{c}_t$ and end-effector (EEF) poses $\mathbf{a}_t = [x_t, y_t, z_t, r_t, p_t, \psi_t, e_t]$ for each timestep $t \in [t_1, t_L]$, where $[x_t, y_t, z_t]$ denotes the 3D Cartesian position, $[r_t, p_t, \psi_t]$ represents the orientation, and $e_t$ indicates the gripper openness. Formally, the model learns to approximate the conditional distribution:
\begin{align} 
    p(x_{t_1:t_L} \mid x_{t_0}, \mathbf{X}_{hist}, \mathbf{a}_{t_1:t_L}, \mathbf{c}_{t_1:t_L})
\end{align}
To achieve consistent long-horizon generation, we employ an autoregressive sliding window where the last predicted frame $x_{t_L}$ is promoted to the new current state $x_{t_0}$ for the next chunk, as illustrated in \autoref{fig:ARgeneration}.

\begin{figure*}[t!] 
    \centering
    \includegraphics[width=0.9\linewidth]{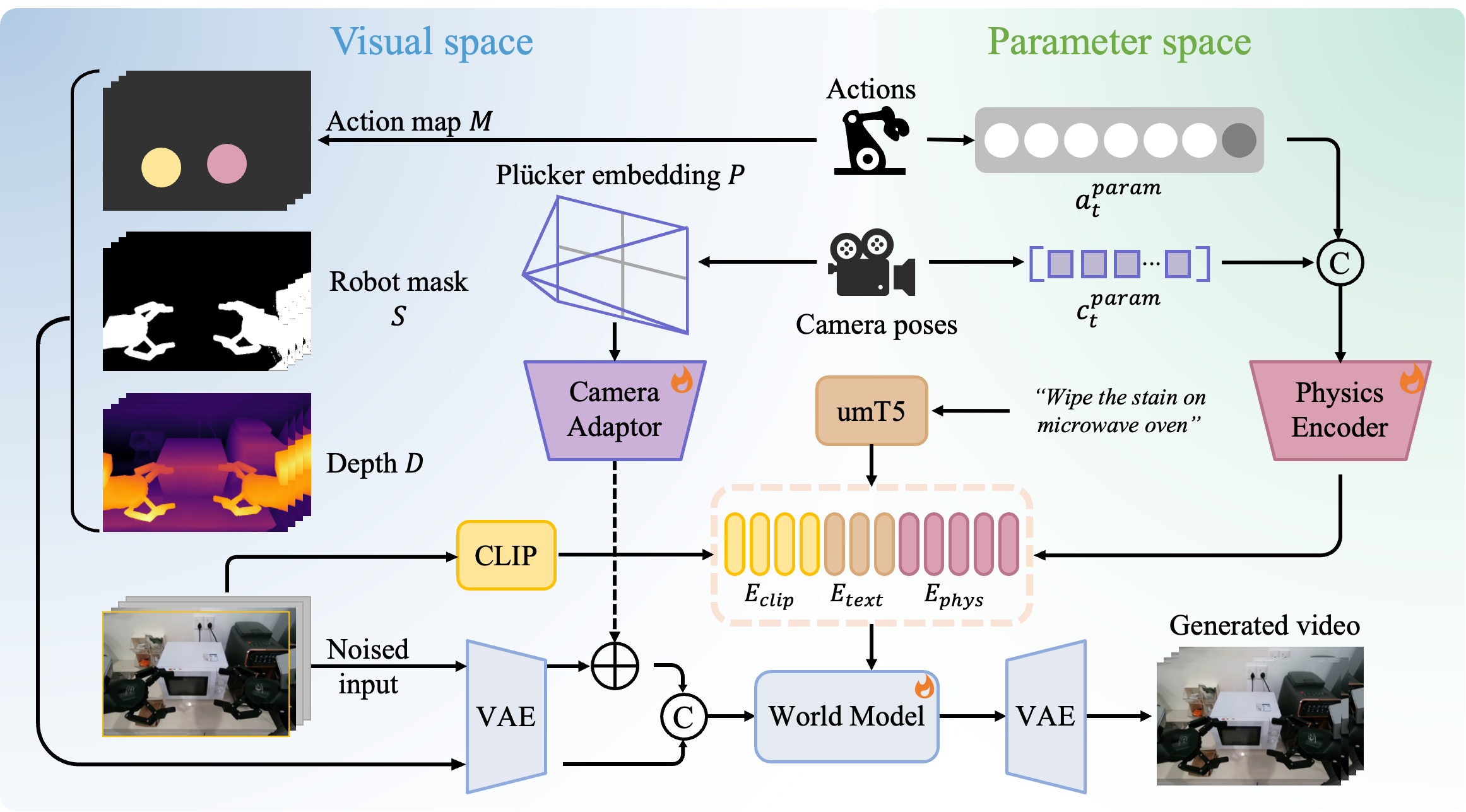}  
    \vspace{-1mm}  
    \caption{\name\ bridges the representation gap between numerical commands and scene dynamics through a dual-space collaboration strategy. (a) Visual Space establish explicit structural anchors by integrating pixel-aligned action map, camera poses, depth and robot masks. (b) Parameter Space preserves the action precision by encoding raw actions and camera matrices into structured numerical embeddings. (c) A dual-path strategy inject both space into generative world model, realizing consistent long-horizon generation.}  
    \vspace{-6mm}
    \label{fig:pipe}
\end{figure*}

\noindent \textbf{Framework Overview.} As illustrated in \autoref{fig:pipe} , \name\ employs a dual-space collaboration strategy to bridge the representation gap. This architecture integrates Visual Space Representations and Parameter Space Representations through a collaborative interaction mechanism, ensuring consistent long-horizon generation that is both spatially grounded and physically driven.

\subsection{Visual Space Representations}

\noindent \textbf{Action.} To bridge the gap between numerical EEF poses and visual synthesis, we transform the raw action sequence $\mathbf{a}_{t_1:t_L}$ into a series of spatial-aware action maps\cite{EnerVerse-AC} $\mathbf{M} \in \mathbb{R}^{L \times 3 \times H \times W}$. This process projects the end-effector's 3D motion into the 2D pixel space, providing explicit guidance for the generative backbone. 

Specifically, for each future timestep $t \in \{t_1, \dots, t_L\}$, the EEF pose in world coordinates $\mathbf{p}_t = [x_t, y_t, z_t]^\top$ is extracted from $\mathbf{a}_t$. Using the calibrated camera intrinsic matrix $\mathbf{K}$ and the time-varying extrinsic parameters $(\mathbf{R}_t, \mathbf{t}_t)$, we transform $\mathbf{p}_t$ into the camera coordinate system $\mathbf{p}_{c,t} = [x_c, y_c, z_c]^\top$ :
\begin{align}
    \mathbf{p}_{c,t} = \mathbf{R}_t \mathbf{p}_t + \mathbf{t}_t
\end{align} 
Subsequently, we project $\mathbf{p}_{c,t}$ onto the image plane using the pinhole camera model to obtain the pixel coordinates $(u_t, v_t)$:
\begin{align}
    u_t = f_x \frac{x_c}{z_c} + c_x, \quad v_t = f_y \frac{y_c}{z_c} + c_y
\end{align}
where $f_x, f_y$ are focal lengths and $(c_x, c_y)$ is the principal point encoded in $\mathbf{K}$. 

To visually represent the future motion, we render a unit circle centered at the projected coordinates $(u_t, v_t)$ for each robotic arm. These renderings from multiple timesteps are stacked into a trajectory video to maintain the temporal continuity. Moreover, in dual-arm scenarios, the trajectories for the left and right arms are differentiated using distinct color schemes to prevent semantic ambiguity. These action maps are rendered on a neutral background to isolate the motion priors, providing the generative model with intuitive spatial constraints. The unit circle uses a 50-pixel radius, empirically chosen to cover the main EEF region at 320$\times$512 resolution.

\noindent \textbf{Camera Pose.} To provide a more informative and spatially-aware description of the observation view, we represent the camera pose for each pixel $(u_t, v_t)$ using Plücker embeddings. Unlike raw extrinsic and intrinsic matrices which are globally defined, Plücker embeddings provide a localized geometric interpretation for every pixel, which is more compatible with the spatial inductive biases of generative video models.

Specifically, for a pixel at coordinates $(u_t, v_t)$ at timestep $t \in \{t_1, \dots, t_L\}$, the Plücker embedding is defined as a $6$-dimensional vector $\mathbf{p}_{u_t,v_t,t} = (\mathbf{o}_t \times \mathbf{d}_{u_t,v_t,t}, \mathbf{d}_{u_t,v_t,t}) \in \mathbb{R}^6$. Here, $\mathbf{o}_t \in \mathbb{R}^3$ denotes the camera center in world coordinate space at time $t$, and $\mathbf{d}_{u_t,v_t,t} \in \mathbb{R}^3$ represents the unit direction vector pointing from the camera center to the pixel $(u_t, v_t)$. The direction vector $\mathbf{d}_{u_t,v_t,t}$ is calculated through the inverse projection process:
\begin{align}
    \mathbf{d}_{u_t,v_t,t} = \text{Normalize}\left( \mathbf{R}_t \mathbf{K}^{-1} [u_t, v_t, 1]^\top \right)
\end{align}
where $\mathbf{R}_t$ is the camera-to-world rotation matrix and $\mathbf{K}$ is the intrinsic matrix. The camera center $\mathbf{o}_t$ is directly derived from the translation vector $\mathbf{t}_t$.

For each frame $t$ in the target chunk, we construct a spatial embedding map $\mathbf{P}_t \in \mathbb{R}^{6 \times H \times W}$ by computing the Plücker coordinates for all pixels. The sequence of spatial observation directions for the $L$-frame chunk is thus represented as $\mathbf{P} \in \mathbb{R}^{L \times 6 \times H \times W}$. By transforming numerical pose parameters into a pixel-wise spatial structure, Plücker embeddings enable the model to better leverage its spatial geometric priors, ensuring the generated video sequences strictly adhere to the intended observation directions.

\noindent \textbf{Scene Geometry and Embodiment Morphology.} 
To complement motion-centric trajectory and camera priors, the Visual Space further incorporates explicit 3D scene geometry and robotic morphological grounding. These components are essential for perceiving spatial occupancy and embodiment–environment relationships, thereby mitigating geometric inconsistencies such as spatial violations and self-intersections. For each $L$-frame chunk, we pre-extract frame-aligned depth and mask sequences to provide dense spatial grounding for visual generation.

For scene geometry, we adopt Video Depth Anything~\cite{VDA} to generate relative depth sequences, denoted as $\mathbf{D} \in \mathbb{R}^{L \times 1 \times H \times W}$. These sequences endow the model with a persistent spatial layout and object-distance priors, ensuring that generated interactions remain consistent with the underlying 3D environmental geometry.

For embodiment morphology, we deploy an automated segmentation pipeline to produce robotic morphological masks, denoted as $\mathbf{S} \in \mathbb{R}^{L \times 1 \times H \times W}$, which strengthens joint-level and body-shape grounding. Specifically, we combine a fine-tuned robotic arm detection model with SAM2~\cite{SAM2} to achieve accurate segmentation of the robot embodiment. To ensure reliable segmentation tracking and handle intermittent visibility (e.g., the robotic arm moving in and out of the frame), we adopt a dynamic template update strategy that iteratively refreshes reference features within the segmentation pipeline. Detailed implementations are provided in Appendix A.

By integrating these geometric and morphological groundings, the Visual Space establishes comprehensive structured spatial priors for generation. This design guarantees that the generated visual states are not only action-driven but also constrained by scene spatial configurations and the inherent kinematic limits of the robot embodiment.

\subsection{Parameter Space Representations}

\noindent \textbf{Camera Pose.} For the camera extrinsic parameters, we adopt a flattening operation~\cite{MotionCtrl} to process the rotation-translation matrix $[R_t | T_t]$ at each timestep $t \in \{t_1, \dots, t_L\}$. The $3 \times 3$ rotation matrix and $3 \times 1$ translation vector are sequentially flattened into a 12-dimensional numerical vector $\mathbf{c}_t^{\text{param}} \in \mathbb{R}^{12}$. This vector fully encapsulates the camera's global spatial transformation, providing a rigorous numerical cue to guide view synthesis.

\noindent \textbf{Dual-Arm Action.} To ensure the numerical stability of orientation representations, we convert Euler angles $[r_t, p_t, \psi_t]$ from raw action $\mathbf{a}_t$ into a four-dimensional unit quaternion $\mathbf{q}_t = [q_x, q_y, q_z, q_w]_t$. A single robotic arm's state is thus re-parameterized as an 8-dimensional vector $[x, y, z, q_x, q_y, q_z, q_w, o]_t \in \mathbb{R}^8$, where $o$ denotes the gripper state. For dual-arm configurations, the feature vectors of both arms are concatenated to form a 16-dimensional \textbf{comprehensive} action feature $\mathbf{a}_t^{\text{param}} \in \mathbb{R}^{16}$.

\noindent \textbf{Structured Physical Embedding.} For each frame in the target chunk, we concatenate the action and camera features into a unified physical command vector. This sequence is then fed into a lightweight Physics Encoder, comprising stacked linear layers and non-linear activations, which maps the numerical features into the latent space of the generative model. The derivation of the physics embedding from the raw control signals is formulated as:
\begin{equation}
    \mathbf{E}_{\text{phys}} = \text{Encoder}_{\text{phys}}\left(\left\{\mathbf{a}_t^{\text{param}} \oplus \mathbf{c}_t^{\text{param}}\right\}_{t=t_1}^{t_L}\right)
\end{equation}
The resulting embeddings $\mathbf{E}_{\text{phys}}$ retain the precise quantitative information of the original control signals. They are formatted for cross-space collaborative interaction, serving as a reliable numerical driver to ensure accurate motion execution during the generative process.

\subsection{Collaborative Interaction Mechanism}

\noindent \textbf{Geometric Grounding via Dense Structural Priors.} Within the Visual Space, we establish a pixel-aligned grounding mechanism to define the framework for generation. The RGB input frames, alongside the action maps $\mathbf{M}$, depth maps $\mathbf{D}$, and morphological masks $\mathbf{S}$, are individually encoded into the latent space using a pre-trained VAE, yielding $\mathbf{z}_{\text{rgb}}$, $\mathbf{z}_{\text{act}}$, $\mathbf{z}_{\text{dep}}$, and $\mathbf{z}_{\text{mask}}$. To maintain causality and prevent information leakage during the synthesis of future $L$ frames, the depth and mask sequences for these frames are populated with values from the current frame $x_{t_0}$, serving as a steady geometric anchor and positional bias. Simultaneously, the Plücker embeddings $\mathbf{P}$ are processed through a Camera Adapter---comprising multi-stage downsampling convolutions and residual blocks---to achieve alignment with the latent dimensions. To ensure computational efficiency and convergence stability in Diffusion Transformer (DiT) backbones, we utilize a $1 \times 1$ convolutional Reducer to compress the redundant features of the auxiliary maps. The final fused latent $\mathbf{z}_{\text{in}}$ is constructed by integrating camera perspectives via element-wise addition and concatenating the reduced spatial priors along the channel dimension:
\begin{equation}
\begin{aligned}
    \mathbf{z}_{\text{spatial}} &= \text{Red.}(\mathbf{z}_{\text{act}}) \oplus \text{Red.}(\mathbf{z}_{\text{dep}}) \oplus \text{Red.}(\mathbf{z}_{\text{mask}}) \\
    \mathbf{z}_{\text{in}} &= [\mathbf{z}_{\text{rgb}} + \text{Adapter}(\mathbf{P})] \oplus \mathbf{z}_{\text{spatial}}
\end{aligned}
\end{equation}
This fused latent $\mathbf{z}_{\text{in}}$ serves as the primary input to the denoising backbone, providing a geometric and morphological template to guide the generative process.

\noindent \textbf{Physical Modulation via Precise Numerical Drivers.} While the Visual Space provides the necessary geometric anchoring, the Parameter Space functions as the essential numerical driver for physical execution. The physics embedding $\mathbf{E}_{\text{phys}}$ is integrated with the global visual features of a reference image $\mathbf{E}_{\text{clip}}$ and, where applicable, the textual instructions $\mathbf{E}_{\text{text}}$ to form a unified semantic sequence:
\begin{align}
    \mathbf{C}_{\text{global}} = \left[\mathbf{E}_{\text{phys}} \oplus \mathbf{E}_{\text{clip}} \oplus \mathbf{E}_{\text{text}}\right]
\end{align}
This consolidated representation $\mathbf{C}_{\text{global}}$ is injected into the generative backbone via cross-attention layers. In this interaction, the latent spatial features function as the query, while $\mathbf{C}_{\text{global}}$ serves as the key and value. Such a mechanism ensures that the synthesized motion is not only spatially consistent within the established scene geometry but also tethered to the underlying numerical commands. This dual-source modulation provides a high-fidelity solution for coordinating complex robot-object interactions.

\section{Experiments}

\subsection{Experimental Setup}

\noindent \textbf{Datasets.} We evaluate \name\ on the AgiBotWorld-Beta dataset~\cite{Agibotworld}, selecting 10 representative tasks that span diverse motion primitives and encompass both rigid and deformable object interactions. Detailed descriptions are provided in Appendix B. The training set (1,000 trajectories, 100 per task) and the disjoint test set (94 unseen clips, 3--45s at 30 FPS) are constructed via randomized sampling to ensure objectivity. This rigorous partitioning facilitates a fair assessment of model stability and robustness during long-horizon generation.

\noindent \textbf{Implementation.} To demonstrate the backbone-agnostic nature of \name, we integrate it into two distinct architectures:
\begin{itemize}
    \item \textbf{UNet-based:} We fine-tune DynamiCrafter~\cite{dynamicrafter} for 40K iterations (per-GPU batch size of 8, learning rate of $5\times 10^{-5}$) using the standard noise prediction loss:
    \begin{equation}
    \mathcal{L}_{\text{UNet}} = \mathbb{E}_{\mathbf{z}_0, \epsilon, t} \left[ \left\| \epsilon - \epsilon_\theta(\mathbf{z}_t, t, \mathbf{C}_{\text{global}}) \right\|_2^2 \right]
    \end{equation}
    where $\mathbf{z}_0$ denotes the clean spatial latent and $\mathbf{z}_t$ is its noisy counterpart.
    \item \textbf{DiT-based:} We utilize the Wan2.2-TI2V-5B model, trained for 20K iterations (per-GPU batch size of 1, learning rate of $2\times 10^{-5}$) with a flow-matching objective:
    \begin{equation}
    \mathcal{L}_{\text{DiT}} = \mathbb{E}_{\mathbf{z}_0, \mathbf{z}_1, t} \left\| u(\mathbf{z}_t, t, \mathbf{C}_{\text{global}}) - \mathbf{v}_t \right\|_2^2
    \end{equation}
    where $\mathbf{z}_1 = \mathbf{z}_{\text{spatial}}$, $\mathbf{z}_0 \sim \mathcal{N}(0, \mathbf{I})$, and $\mathbf{C}_{\text{global}}$ incorporates textual embeddings.
\end{itemize}

Models are trained on 8$\times$80GB GPUs, while CLIP, VAE, and T5 encoders remain frozen. Each generation chunk predicts $L=16$ frames at $320 \times 512$ resolution.

\noindent \textbf{Temporal Conditioning \& Action Formulation.} We employ a sparse sampling strategy: $N_{hist}$ frames are randomly sampled from preceding sequences during training and uniformly from generated chunks during inference to capture long-range context efficiently. Action representation follows~\cite{EnerVerse-AC} by using absolute end-effector positions to ensure precise spatial alignment and fair comparison.

\noindent \textbf{Inference Efficiency \& Pre-processing.} To address potential concerns regarding the computational overhead of visual priors (e.g., mask and depth extraction), we adopt an optimized pipeline. During training, all geometric and semantic features are pre-computed offline to maximize throughput. For online inference, feature extraction is performed on-the-fly only for the sparse reference set ($N_{hist}$ historical frames and 1 initial frame per chunk). This localized extraction ensures that the pre-processing remains a minor fraction of the total latency. Consequently, \name\ achieves an end-to-end inference speed of 1.918s per chunk (including feature extraction and forward pass), demonstrating suitability for high-frequency reactive control without incurring significant temporal bottlenecks.

\noindent \textbf{Evaluation Metrics.} We report standard fidelity (PSNR, SSIM, LPIPS) and the EWMBench~\cite{EWMBench} framework, which assesses: (1) Motion Correctness: EEF trajectory adherence via YOLO-World~\cite{YOLO-World} tracking (HSD, NDTW, DYN); (2) Semantic Alignment: instruction-following via MLLM-based scores; and (3) Scene Consistency: geometric stability via DINOv2 similarity. All EWMBench scores are normalized to $[0, 1]$. Final results are averaged over 15 evaluations (5 post-convergence checkpoints $\times$ 3 random seeds). Detailed metric definitions are in Appendix B.

\begin{figure*}[htbp] 
    \centering
    \includegraphics[width=0.95\linewidth]{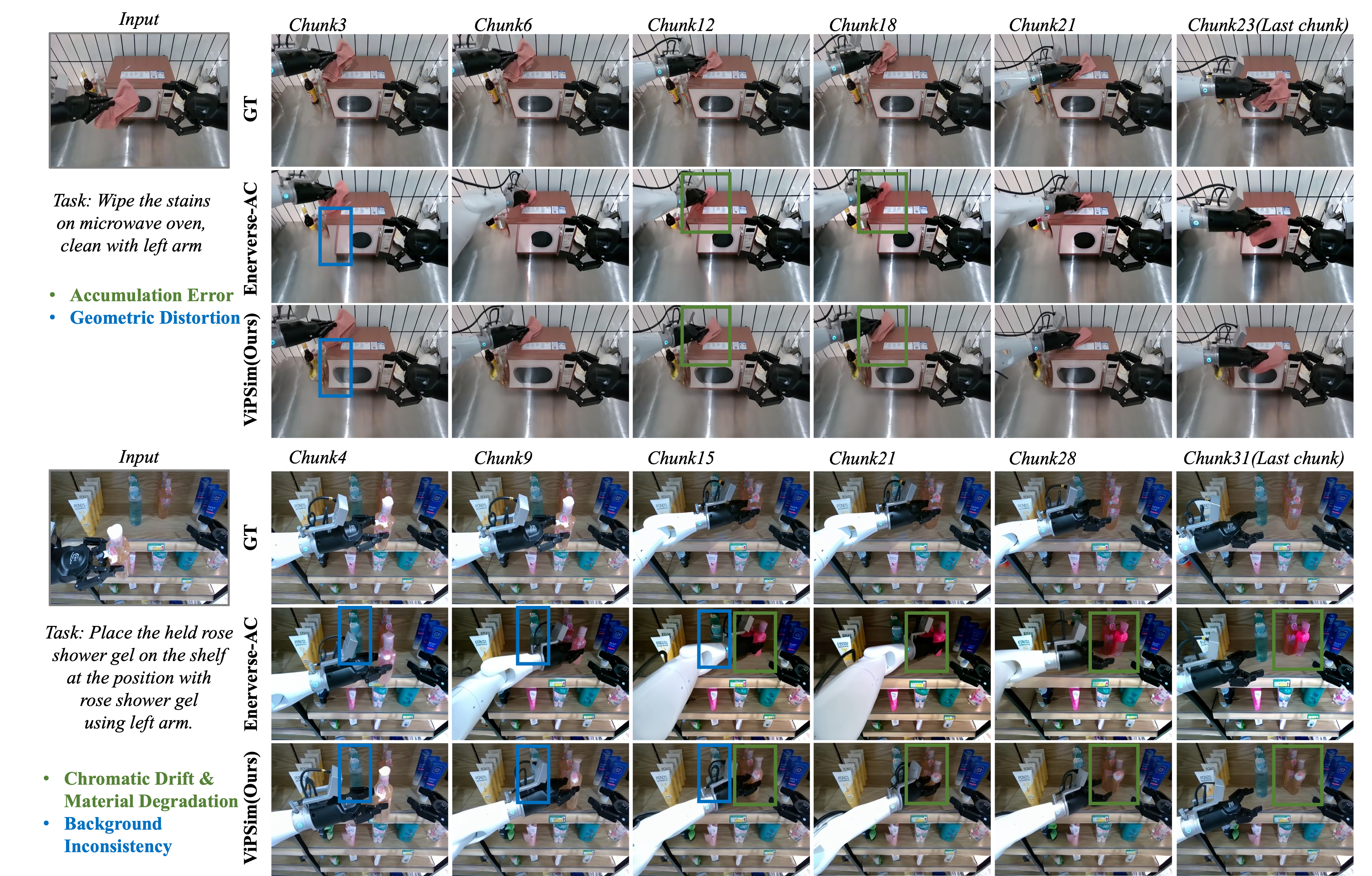}  
    \vspace{-1mm}  
    \caption{Qualitative comparison of long-horizon temporal stability. We evaluate ViPSim against the baseline on tasks requiring extended consistency: Wiping the table (top) and Shelving shampoo (bottom). Green bounding boxes denote the structural integrity of manipulated objects under continuous manipulation, while blue bounding boxes highlight the model's ability to maintain and recover environmental context after occlusion by the robotic arm.} 
    \vspace{-6mm}
    \label{fig:res2}
\end{figure*}

\subsection{Quantitative Analysis}

As shown in \autoref{table1-1}, quantitative evaluations for long-trajectory generation demonstrate that \name(UNet) consistently outperforms the EnerVerse-AC baseline across all perceptual dimensions. Notably, \name(DiT) achieves a significant performance leap, attaining optimal values for PSNR, SSIM, and LPIPS. To ensure statistical rigor, we report the mean and standard deviation across 15 independent evaluations (5 consecutive checkpoints $\times$ 3 random seeds) in \autoref{table1-3}, which confirms the stability of our generative framework.

We further conduct a comparative evaluation following the EWMBench protocol. As summarized in \autoref{table1-2}, \name\ yields consistent improvements across all metrics. Specifically, our framework manifests a commanding lead in motion correctness by establishing rigorous motion-visual synchronization, ensuring that generated sequences adhere to input actions. Building upon \name(UNet), the \name(DiT) variant further pushes these boundaries, effectively narrowing the gap between generated trajectories and real-world dynamics. Beyond motion fidelity, \name\ excels in semantic alignment and scene consistency, demonstrating a robust capacity for maintaining spatiotemporal coherence over extended horizons. These results underscore the efficacy of \name\ in delivering motion-consistent, high-fidelity rollouts. Detailed task-specific evaluations are provided in Appendix C.

\begin{table}[htbp]
\centering
\footnotesize 
\caption{Quantitative results for long-trajectory generation on the constructed AgiBotWorld-Beta dataset.}
\setlength{\tabcolsep}{3pt}
\begin{tabular}{lccc}
\toprule
\multirow{2}{*}{Model} & \multicolumn{3}{c}{Quality}  \\
\cmidrule(r){2-4}
& PSNR $\uparrow$& SSIM $\uparrow$& LPIPS $\downarrow$\\
\midrule
 EnerVerse-AC~\cite{EnerVerse-AC}  & 17.93&0.73&0.25\\
 ViPSim(UNet) &18.42&0.74&0.23  \\
 ViPSim(DiT)& \textbf{20.35} & \textbf{0.80} & \textbf{0.19} \\
\bottomrule
\end{tabular}
\vspace{-2mm}
\label{table1-1}
\end{table}

\begin{table}[htbp]
\centering
\caption{Statistical reliability analysis of \name(DiT) over 15 independent runs.}
\label{table1-3}
\vspace{-5pt}
\resizebox{\linewidth}{!}{
\begin{tabular}{ccccccc}
\toprule
Model& DYN $\uparrow$ & HSD $\uparrow$ & nDTW $\uparrow$ & PSNR $\uparrow$ & SSIM $\uparrow$ & LPIPS $\downarrow$ \\
\midrule
\makecell{ViPSim(DiT)}  & \makecell{0.7534 \\ $\pm$ 0.0184} & \makecell{0.8383 \\ $\pm$ 0.0172} & \makecell{0.8836 \\ $\pm$ 0.0183} & \makecell{20.3543 \\ $\pm$ 0.0831} & \makecell{0.8013\\ $\pm$ 0.0024} & \makecell{0.1920 \\ $\pm$ 0.0044} \\
\bottomrule
\end{tabular}
}
\vspace{-6mm}
\end{table}

\subsection{Qualitative Results}

\noindent \textbf{Visual Persistence and Scene Consistency.} To evaluate temporal stability, we visualize long-horizon rollouts to assess the model's resistance to recursive error accumulation. As shown in \autoref{fig:res2}, \name\ maintains remarkable structural and semantic consistency. 

Specifically, the green bounding boxes highlight \name(DiT)'s ability to preserve object identity over prolonged durations. While the baseline suffers from chromatic drift and material degradation—e.g., the shampoo bottle's gel-like texture morphs into an opaque red solid—\name\ successfully anchors photometric and semantic attributes via dense geometric priors.

\begin{table*}[htbp]
\centering
\footnotesize 
\caption{Comparative evaluation of ViPSim against baseline world models, assessing motion correctness, semantic alignment, scene consistency, and diversity via the EWMBench~\cite{EWMBench} protocol. }
\begin{tabular}{l|ccccccccc|c|c}
\toprule
\multirow{2}{*}{Model} & \multicolumn{4}{c}{Motion $\uparrow$} & \multicolumn{5}{c}{Semantics $\uparrow$} & Scene $\uparrow$ & \multirow{2}{*}{Overall $\uparrow$} \\
\cmidrule(r){2-5} \cmidrule(r){6-10} \cmidrule(r){11-11}
& Dyn & Hsd & nDTW & Motion Sum. & BLEU & CLIP & Diversity & Logics & Semantic Sum. & SceneC & \\
\midrule
EnerVerse-AC~\cite{EnerVerse-AC}  & 0.5723 & 0.6979 & 0.7802 & 2.0504 & 0.2835 & 0.9038 & \textbf{0.0092} & 0.8839 & 2.0804 & 0.9078 & 5.0386 \\
ViPSim(UNet) & 0.7029 & 0.8279 & 0.8721 & 2.4030 & 0.2864 & 0.9057 & 0.0091 & 0.9140 & 2.1151 & 0.9111 & 5.4291 \\
ViPSim(DiT) & \textbf{0.7534} & \textbf{0.8383} & \textbf{0.8836} & \textbf{2.4754} & \textbf{0.3247} & \textbf{0.9125} & 0.0085 & \textbf{0.9333} & \textbf{2.1790} & \textbf{0.9153} & \textbf{5.5697} \\
\bottomrule
\end{tabular}
\vspace{0.5em}
\label{table1-2}
\end{table*}

\begin{figure*}[t!] 
    \centering
    \includegraphics[width=1.0\linewidth]{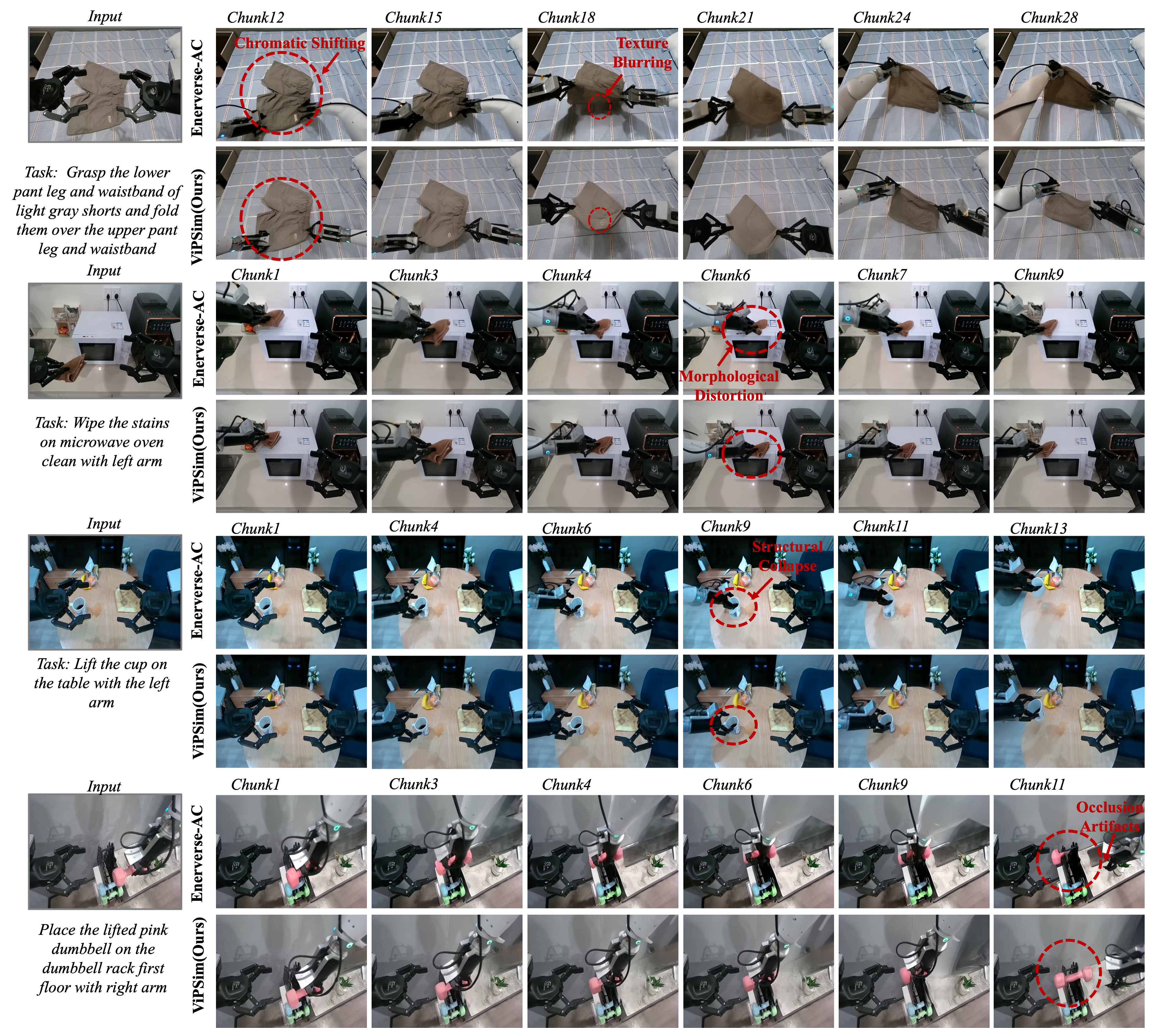}  
    \vspace{-5mm}  
    \caption{Qualitative comparison of deformable and rigid object manipulation  between ViPSim and baseline methods. The first four rows illustrate interactions with deformable objects (e.g., clothes folding and wiping). The latter four rows show rigid object manipulation (e.g.,lift the cup and dumbbell), red circle provide magnified views of key interaction areas.}  
    \vspace{-6mm}
    \label{fig:res1}
\end{figure*}

Simultaneously, the blue bounding boxes demonstrate our model's proficiency in restoring environmental elements previously occluded by the robotic arm. Unlike the baseline, where background components (e.g., shelf items) undergo geometric distortion upon disocclusion, \name\ consistently retains intricate scene details, exhibiting superior spatiotemporal anchoring.

\begin{table}[htbp]
\centering
\footnotesize 
\setlength{\tabcolsep}{3pt}
\caption{Quantitative comparison of different variants on perceptual quality, with performance measured via PSNR, SSIM and LPIPS.}
\begin{tabular}{cccc|cc|ccc}
\toprule
 \multicolumn{4}{c|}{Visual Space} & \multicolumn{2}{c|}{Para Space} & \multicolumn{3}{c}{Quality}  \\
\cmidrule(r){1-4} \cmidrule(r){5-6} \cmidrule(r){7-9}

\makecell{action\\map} & plk & depth & mask &  \makecell{raw\\action} & \makecell{concat\\RT} &\makecell{PSNR\\$\uparrow$} & \makecell{SSIM\\$\uparrow$} & \makecell{LPIPS\\$\downarrow$} \\
\midrule
\ding{51} &  &  &  &  &  & 17.9266 & 0.7336 & 0.2571 \\
\ding{51} &  &  &  &   \ding{51}  &  & 18.1805&0.7315& 	0.2489 \\
\ding{51}  & \ding{51}   &  &  &\ding{51} &\ding{51} & 18.3953 & 0.7330 & 0.2459 \\
\midrule
\ding{51}&  & \ding{51} &  &  &  & 18.1970 & 0.7399 & 0.2539\\
\ding{51}&  &  & \ding{51} &&& 18.1168 & 0.7388 & 0.2518 \\
\ding{51}&  &\ding{51}&\ding{51}&  &  & 18.4183&\textbf{0.7421}& 0.2442  \\
\midrule
\ding{51}&\ding{51}&\ding{51}&\ding{51}&\ding{51}&\ding{51}& \textbf{18.4231} & 0.7392 & \textbf{0.2356} \\
\bottomrule
\end{tabular}
\vspace{-4mm}
\label{table3-1}
\end{table}

We provide a qualitative comparison between \name\ and baseline methods across four representative tasks to evaluate the visual grounding of the generated states. As illustrated in \autoref{fig:res1}, \name\ maintains superior consistency across two interaction types:

\begin{table*}[t]
\centering
\footnotesize 
\caption{Quantitative comparison of different variants with respect to motion correctness, scene consistency, semantic and diversity via the EWMBench~\cite{EWMBench} protocol. Overall score denotes the sum over all EWMBench metrics}
\resizebox{1\textwidth}{!}{
\begin{tabular}{cccc|cc|cccc|ccccc|c|c}
\toprule
\multicolumn{4}{c|}{Visual Space} & \multicolumn{2}{c|}{Para Space} & \multicolumn{4}{c}{Motion $\uparrow$} & \multicolumn{5}{c}{Semantics $\uparrow$} & Scene $\uparrow$ & \multirow{2}{*}{Overall $\uparrow$} \\
\cmidrule(r){1-4} \cmidrule(r){5-6} \cmidrule(r){7-10} \cmidrule(r){11-15} \cmidrule(r){16-16}
\makecell{action\\map} & plk & depth & mask & \makecell{raw\\action} & \makecell{concat\\RT} & Dyn & Hsd & nDTW & \makecell{Motion\\Sum} & BLEU & CLIP & Diversity& Logics & \makecell{Semantic\\Sum}. & SceneC & \\
\midrule
\ding{51} &  &  &  &  &  &0.5723&0.6979& 0.7802 &2.0504 &0.2835 &0.9038 &0.0092 &0.8839 &2.0804 &0.9078 &5.0386 \\ 
\ding{51} &  &  &  &   \ding{51}  &  &0.6308 &	0.8007 	&0.8491 &	2.2807& 0.2835& 0.9053 &0.0095&0.9032 &2.1015&0.9093& 5.2915  \\ 
\ding{51}  & \ding{51}   &  &  &\ding{51} &\ding{51} &0.6646 &	0.7823 &0.8446 &2.2915 &0.2837 &0.9043 &\textbf{0.0099} &\textbf{0.9333} &\textbf{2.1313} &0.9086 &5.3314  \\ 
\midrule
\ding{51}&  & \ding{51} &  &  &  &0.5730&0.7289&0.8016&2.1036&0.2914&0.9062&0.0091&0.8667&2.0734&0.9077&5.0847 \\ 
\ding{51} &  &  &   \ding{51}&   &  &0.5716 &0.7165 &0.7962 &2.0843 &\textbf{0.2942} &\textbf{0.9077} &0.0091 &0.8903 &2.0415 &0.9090 &5.0348 \\ 
\ding{51}&  &\ding{51}&\ding{51}&  &  & 0.6121 &0.7716 &0.8460 &2.2297 &0.2803 &0.9016 &0.0077 &0.9183 &2.1079 &0.9109 &5.2485 \\ 
\midrule
\ding{51}&\ding{51}&\ding{51}&\ding{51}&\ding{51}&\ding{51}& \textbf{0.7029}&\textbf{0.8279}&\textbf{0.8721}&\textbf{2.4030}&0.2864&0.9057&0.0091&0.9140&2.1151&\textbf{0.9111}&\textbf{5.4291} \\ 
\bottomrule
\end{tabular}
\vspace{0.5em}
}
\vspace{-4mm}
\label{table3-2}
\end{table*}

\begin{figure*}[t] 
    \centering
    \includegraphics[width=0.95\linewidth]{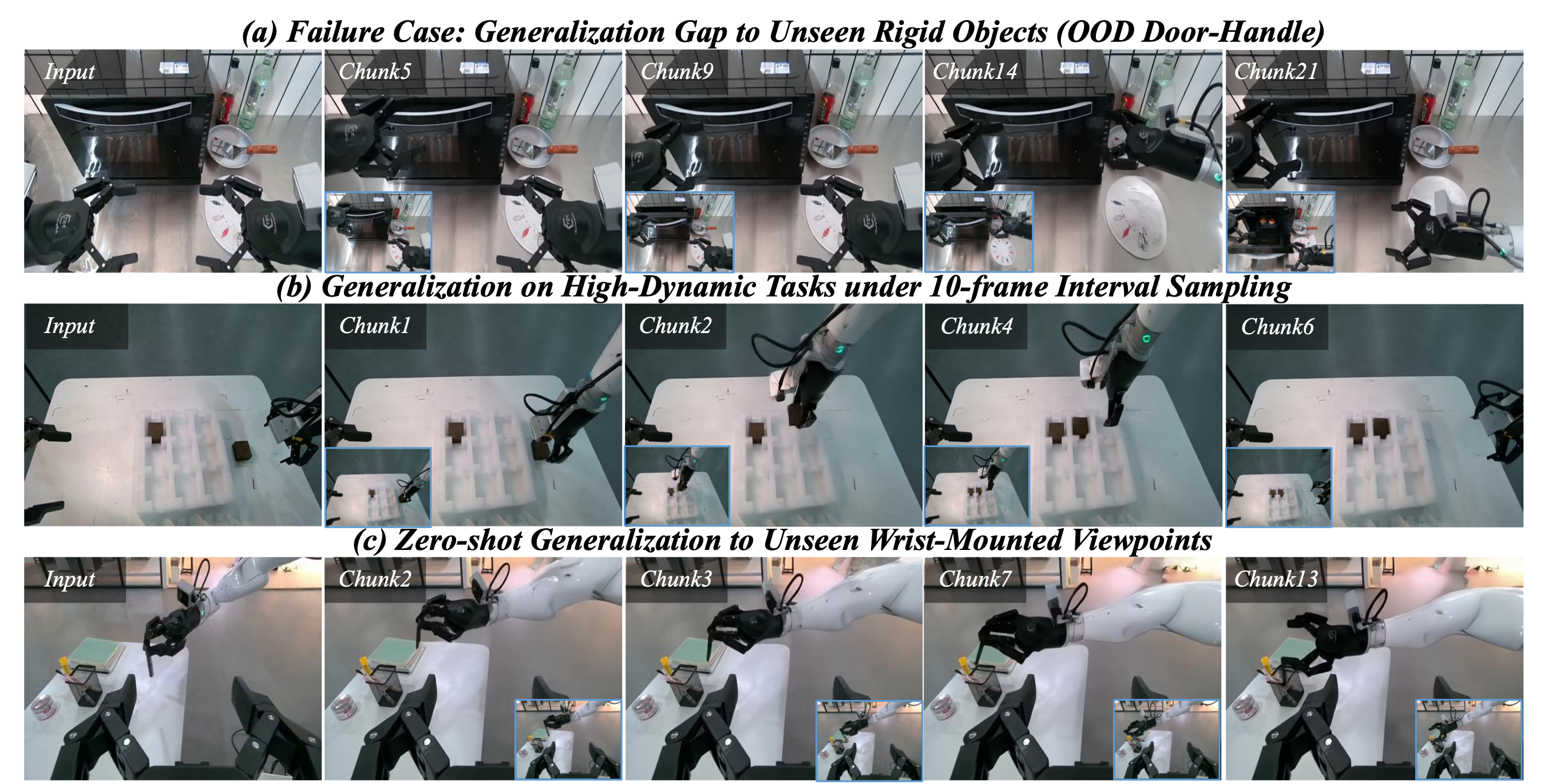}  
    \vspace{-1mm}  
    \caption{Analysis of dynamic compatibility, robustness, and generalization limits.
    (a) Unseen task generalization. (b) High-dynamic stress test. (c) Unseen view generalization. }
    \vspace{-6mm}
    \label{fig:rebuttal_wgt}
\end{figure*}

\noindent \textbf{Interaction with Deformable Objects.} 

In tasks like \textit{Folding Shorts} and \textit{Wiping the Stain}, baselines exhibit progressive textural degradation and morphological warping. For garment manipulation, the baseline suffers total loss of surface detail by Chunk 18, whereas \name\ maintains exceptional fabric fidelity. In the wiping task, \name\ accurately renders fine-grained folding gaps, preserving the structural coherence of non-rigid bodies.

\noindent \textbf{Interaction with Rigid Objects.} 

In tasks such as \textit{Pick up a Cup} and \textit{Release a Dumbbell}, \name\ ensures superior object permanence. While interaction with the gripper triggers structural collapse in the baseline (e.g., near Chunk 9 in the cup task), \name\ preserves both global geometry and fine surface details, even under complex occlusions.

\noindent \textbf{Dynamic Compatibility.} 
To test robustness, we conduct high-dynamic stress tests by subsampling end-effector poses at a 10-frame interval. Despite severe inter-frame jumps, \name\ infers stable robot motion aligned with the ground truth (\autoref{fig:rebuttal_wgt}b), proving the efficacy of our dual-space collaboration under sparse motion constraints. 

\begin{figure}[t!]  
    \centering
    \includegraphics[width=\linewidth]{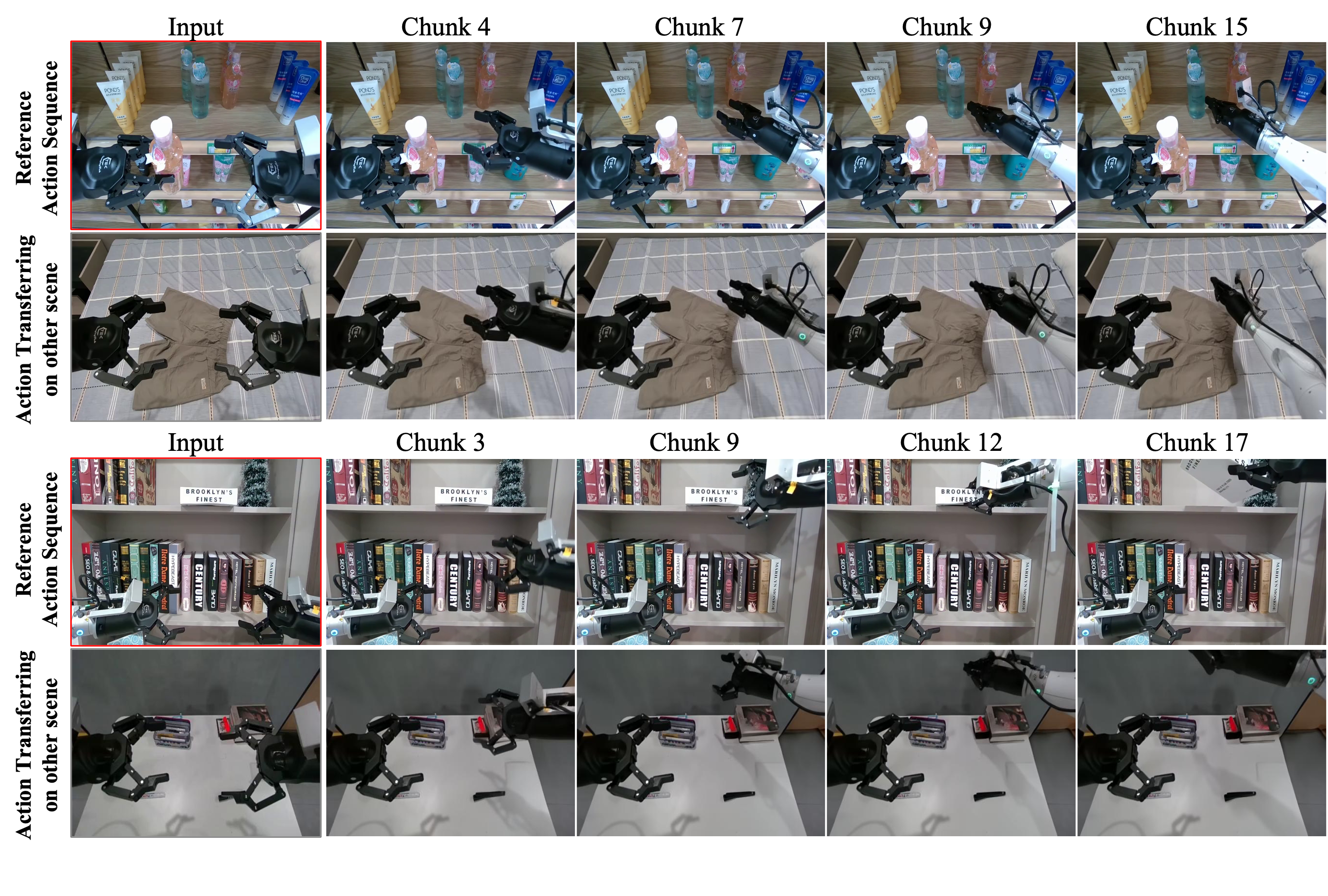}  
    \vspace{-4mm}
    \caption{Evaluation of action compliance through swapping. The action from a source sequence (top row) is transplanted into a different scene. \name\ demonstrates precise adherence to the transplanted actions, proving that the generated motion is driven by numerical action.}
    \vspace{-6mm}
    \label{fig:action-follow}
\end{figure}

\noindent \textbf{Action Faithfulness.} Cross-action swap experiments (\autoref{fig:action-follow}) demonstrate that \name\ precisely follows reference trajectories regardless of the initial scene context. This confirms the efficacy of our parameter-space embeddings in steering the generative process with high quantitative precision.

\subsection{Ablation Study}
We conduct a systematic ablation study to evaluate the individual contributions of components within the Visual and Parameter Spaces. Quantitative results are summarized in \autoref{table3-1} and \autoref{table3-2}. Our analysis focuses on how the integration of granular spatial and parametric cues collectively bridges the representation gap in long-horizon world modeling.

\noindent \textbf{Effectiveness of Dual-Space Collaboration.} The results demonstrate that visual-spatial priors and parametric signals are mutually indispensable. The baseline utilizing only an \textit{Action Map} yields the lowest performance, as pixel-level cues alone lack the numerical grounding necessary for precise motion scaling. Integrating \textit{Raw Action} into the Parameter Space leads to a marked performance increase, providing a critical metric scale that anchors the visual synthesis. Furthermore, while the \textit{Camera-Conditioned} variant enhances structural logic through geometric embeddings, the \textit{Full Strategy}---incorporating fine-grained \textit{Mask Map} and \textit{Depth Map} grounding---achieves the best performance across all metrics. This synergy demonstrates that the intersection of parametric precision and geometry-aware visual grounding is fundamental to generating high-fidelity states in complex tasks.

\noindent \textbf{Impact of Fine-grained Geometric Grounding.} This ablation investigates how incremental geometric information reinforces the structural fidelity of generated states. The combination of \textit{Action Map} and \textit{Depth Map} provides essential spatial depth-awareness, significantly reducing visual distortions compared to the baseline. The subsequent integration of the \textit{Mask Map} enables the model to resolve precise object-background boundaries and handle complex interactions. Finally, the \textit{Double Strategy} configuration, which unites this visual grounding with parametric guidance, achieves the highest scores in motion correctness and scene consistency. This progression underscores that while depth provides the 3D scaffold, the addition of mask-based morphological priors and parametric guidance is essential for maintaining the spatiotemporal integrity of the generated environment.

\noindent \textbf{Motion-Centric Metric Prioritization.} While general generation quality metrics (e.g., PSNR, SSIM) may show marginal differences among different variants, \name\ achieves clear and stable improvements in motion-related metrics (\autoref{table3-2}). Given that motion fidelity is critical for long-horizon embodied evaluation, the synergy of our proposed modules leads to a significant performance boost in trajectory adherence and physical plausibility.

\section{Limitations} 
\noindent Despite the promising results, we acknowledge several generalization constraints inherent to our data-driven framework. The primary limitation of \name\ lies in the decoupling of semantic understanding and physical interaction for out-of-distribution (OOD) objects. 

First, as illustrated in \autoref{fig:rebuttal_wgt}(a), while our framework maintains accurate motion trajectories, it occasionally struggles with complex semantic interactions and occlusion-induced hallucinations. For instance, in an unseen ``stove-opening'' task, the model may generate a correct arm trajectory, yet the door remains static. This suggests that the model lacks implicit object-level affordance priors. Potential remedies include: (i) incorporating fine-grained object-centric segmentation to explicitly model part-level dynamics, or (ii) leveraging common-sense world knowledge from large-scale video foundation models as an initialization prior.

Second, we observe morphological deviations under novel viewpoints. As shown in \autoref{fig:rebuttal_wgt}(c), although end-effector trajectories remain precisely aligned with the ground truth, the synthesized robot arm may exhibit geometric artifacts due to OOD camera-to-robot spatial transformations. 

Finally, while the current validation primarily utilizes a static camera setup, our architecture is inherently compatible with dynamic camera poses and multi-view scenarios. The camera-conditioning module within our Parameter Space is designed to support moving viewpoints, including static ego-centric and dynamic wrist-mounted views. We plan to evaluate these capabilities in future iterations to further enhance the system's spatial versatility.

\section{Conclusion} 
\label{sec:conclusion}

In this paper, we presented \name, a novel framework designed to bridge the representation gap between high-dimensional visual synthesis and low-dimensional actions in Embodied World Models by introducing a synergistic collaboration between Visual and Parameter Spaces. Extensive experiments demonstrate that \name\ effectively mitigates cumulative trajectory drift in long-horizon rollouts and maintains high structural consistency even dealing with deformable objects. Furthermore, \name's ability to generalize across out-of-distribution and cross-embodiment scenarios highlights its potential as a robust foundation for evaluating general-purpose embodied agents. In the future, we plan to integrate VLA policies to achieve a complete and reliable closed-loop evaluation workflow, thereby accelerating the iteration of embodied agents.

\newpage
\renewcommand{\refname}{References} 
\bibliographystyle{unsrt}  
\bibliography{references.bib} 

\end{document}